\documentclass[10pt,twocolumn,letterpaper]{article}

\usepackage[pagenumbers]{cvpr} %

\usepackage[T1]{fontenc}
\usepackage[utf8]{inputenc}
\usepackage{microtype}
\usepackage{nicefrac}
\usepackage{multirow}
\usepackage{array}
\usepackage{colortbl}
\usepackage{bm}
\usepackage{float}
\usepackage{algorithm}
\usepackage{algpseudocode}
\usepackage{overpic}
\usepackage{fancyvrb}
\DefineVerbatimEnvironment{promptbox}{Verbatim}{%
  frame=single,framesep=3pt,rulecolor=\color{black!35},
  fontsize=\scriptsize,baselinestretch=0.95,
  xleftmargin=0pt,xrightmargin=0pt}

\definecolor{MyCustomBlue}{HTML}{150CA6}
\definecolor{tealdark}{rgb}{0.45,0.78,0.75}
\definecolor{teallight}{rgb}{0.82,0.93,0.92}
\definecolor{tableyellow}{rgb}{1.0,1.0,0.7}
\definecolor{pipeblue}{HTML}{4A6984}
\definecolor{pipered}{HTML}{C00000}
\definecolor{pipegreen}{HTML}{73B5A5}

\newcommand{\best}{\cellcolor{tealdark}}
\newcommand{\sbest}{\cellcolor{teallight}}

\newcommand{\xmark}{\ensuremath{\times}}

\newcommand{\MethodName}{{\textsc{FlexPart}}\xspace}
\newcommand{\Ours}{\MethodName}
\newcommand{\Bench}{{\textsc{FlexPartBench}}\xspace}
\newcommand{\bg}{\rm{bg}}

\newcommand{\latent}{\mathbf{z}}
\newcommand{\timestep}{t}
\newcommand{\latentt}{\mathbf{z}_{\timestep}}
\newcommand{\mesh}{\rmM}
\newcommand{\latentzero}{\mathbf{z}_{0}}
\newcommand{\noise}{\boldsymbol{\epsilon}}
\newcommand{\velocity}{\mathbf{v}}
\newcommand{\parameter}{\phi}
\newcommand{\decoder}{\varphi}
\newcommand{\sattr}{\mathbf{s}}
\newcommand{\point}{\rvx}

\newcommand{\res}{N}

\newcommand{\vaeencoder}{\psi}

\newcommand{\voxel}{\rvv}
\newcommand{\tokens}{\rmT}

\definecolor{cvprblue}{rgb}{0.21,0.49,0.74}
\usepackage[
  pagebackref,
  breaklinks,
  colorlinks=true,
  citecolor=MyCustomBlue,
  linkcolor=MyCustomBlue,
  urlcolor=MyCustomBlue,
  pdftitle={FlexPart: Text-Controlled 3D Part Segmentation at Flexible Granularity}
]{hyperref}

\usepackage{amsmath,amsfonts,bm}

\def\Tabref#1{Tab.~\ref{#1}}
\def\Secref#1{Sec.~\ref{#1}}
\def\Appref#1{App.~\ref{#1}}
\def\Figref#1{Fig.~\ref{#1}}
\def\Eqref#1{Eq.~\ref{#1}}

\def\eqref#1{equation~\ref{#1}}

\def\1{\bm{1}}

\def\rvv{{\mathbf{v}}}

\def\rvx{{\mathbf{x}}}

\def\rmM{{\mathbf{M}}}

\def\rmT{{\mathbf{T}}}

\def\rmV{{\mathbf{V}}}

\def\vzero{{\bm{0}}}

\def\mI{{\bm{I}}}

\DeclareMathAlphabet{\mathsfit}{\encodingdefault}{\sfdefault}{m}{sl}
\SetMathAlphabet{\mathsfit}{bold}{\encodingdefault}{\sfdefault}{bx}{n}

\def\gN{{\mathcal{N}}}

\def\gP{{\mathcal{P}}}

\def\gT{{\mathcal{T}}}

\def\sR{{\mathbb{R}}}

\newcommand{\E}{\mathbb{E}}
\newcommand{\Ls}{\mathcal{L}}

\newcommand{\reg}{\lambda}

\def\paperID{*****} %
\def\confName{CVPR}
\def\confYear{2027}

\title{FlexPart: Text-Controlled Semantics-Aware 3D Part Segmentation at Flexible Granularity}

\author{%
Yiqun Zhao\textsuperscript{1,2*}\quad
Jiachen Xu\textsuperscript{2}\quad
Yihao Zhi\textsuperscript{3,2*}\quad
Yunfei Zhao\textsuperscript{2}\quad
Zibo Zhao\textsuperscript{2}\quad
Xintong Han\textsuperscript{2}\quad\\
Chunchao Guo\textsuperscript{2,\dag}\quad
Shenghua Gao\textsuperscript{1,4,\dag}\\[0.35em]
{\normalsize
\textsuperscript{1}The University of Hong Kong\quad
\textsuperscript{2}Tencent Hunyuan}\\
{\normalsize
\textsuperscript{3}The Chinese University of Hong Kong, Shenzhen\quad
\textsuperscript{4}Shenzhen Loop Area Institute}\\
{\normalsize Project page: \url{https://zhao-yiqun.github.io/flexpart/}}
}

\makeatletter
\apptocmd{\@maketitle}{%
  \begin{center}
  \includegraphics[width=\textwidth]{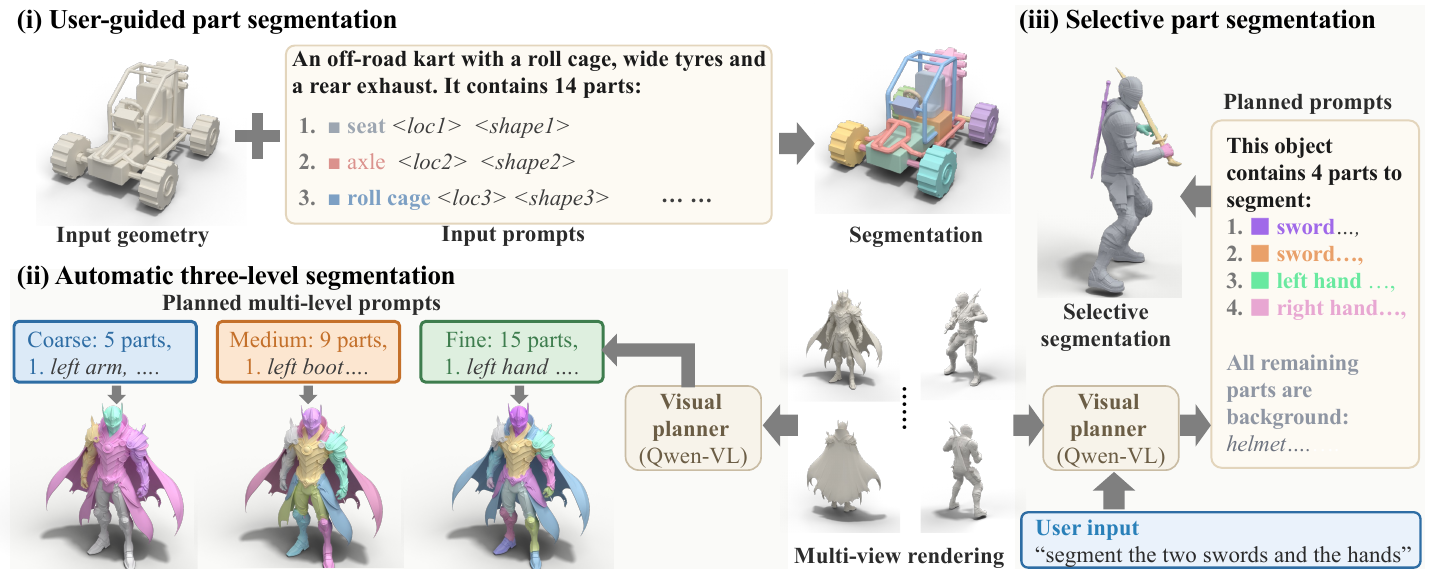}
  \vspace{-2em}
  \captionof{figure}{\textbf{A unified framework for text-driven, semantics-aware 3D part segmentation.} \MethodName{} partitions a given 3D geometry following text prompts. A prompt defines the desired label space using semantic part names augmented with location (<loc>) and shape (<shape>) descriptors. \textbf{(i)} Users can manually author this detailed item list for precise control. \textbf{(ii)} To streamline interaction, a visual planner can automatically construct these prompts from multi-view renderings. By simply varying the planned text prompts, the \MethodName{} model dynamically outputs coarse, medium, or fine partitions. \textbf{(iii)} Given a simple, free-form user command, the planner isolates the requested parts and automatically assigns the remaining parts to the background, enabling selective segmentation.
  }
  \label{fig:teaser}
\end{center}

}{}{\PackageError{main}{Could not patch \string\@maketitle\space to carry the
  teaser. cvpr.sty has changed; place figures/teaser.tex by hand}{}}
\makeatother

\makeatletter
\newcommand\blfootnote[1]{\begingroup\renewcommand\thefootnote{}\footnotetext{#1}\endgroup}
\makeatother

\begin{document}
\maketitle

\iftoggle{cvprfinal}{%
  \blfootnote{\textsuperscript{*}Work done during an internship at Tencent Hunyuan.}%
  \blfootnote{\textsuperscript{\dag}Corresponding authors.}%
}{}

\begin{abstract}

A 3D asset admits multiple valid part decompositions, while semantics-aware segmentation should reflect both the desired granularity and the parts a user wants to extract. Ideally, a practical segmentation model should strive for a simple principle: \textbf{what you ask for is what you get}. However, realizing this goal is hindered by two core obstacles: the lack of multi-granularity training data and the absence of a holistic control mechanism.
We propose \MethodName, a unified framework for text-controlled 3D mesh segmentation. To establish flexible control, we introduce a holistic \emph{control prompt} comprising semantic part names with location and shape clauses. To support this mechanism, we utilize an automated Vision-Language Model (VLM) pipeline to construct a large-scale dataset of multi-granularity partitions and their corresponding prompts. Driven by this data, \MethodName generates a view-independent \emph{part-identity field} in native 3D space, directly assigning discrete part IDs to mesh faces. To compel the model to respect semantic conditions rather than rely on geometric priors, we train the model using randomized background dropping. Furthermore, an inference-time VLM planner can automatically complete descriptive clauses from 2D renderings,  allowing users to simply name the desired parts in ordinary words. \MethodName with
planner-completed prompts outperforms the strongest evaluated baseline by $15\%$ in mIoU, even
when that baseline receives ground-truth guidance. On our held-out \Bench, \MethodName leads at
each granularity and improves average mIoU by $22\%$ over the strongest baseline. With only the
requested part names supplied by the user, it also improves IoU by $35\%$ over the strongest
click-driven baseline on selective segmentation.

\end{abstract}

\section{Introduction}
\label{sec:intro}

3D part segmentation provides explicit boundaries on mesh surfaces, facilitating downstream applications such as part-level editing, rigging, animation, and 3D printing. Crucially, a single 3D asset can be decomposed in multiple valid ways depending on the user's intent: coarse functional groups for retargeting, individual accessories for texturing, or specific named parts for localized edits (as shown in \Figref{fig:teaser}). However, two core obstacles remain in achieving this flexible, intent-driven control. First, existing 3D datasets lack multi-granularity semantic annotations. Second, the absence of a holistic, view-independent control signal often leads to fragmented boundaries and unresolved occlusions. Therefore, developing a unified framework that couples a global semantic control signal with multi-granularity decomposition is essential to realize the user-centric principle: \textbf{what you ask for is what you get}.

Regarding the first obstacle, existing 3D datasets typically provide only a single, fixed partition per geometry, mostly sourced from open-source repositories or fixed artist creations, lacking consistent multi-granularity semantics-aware annotations. Recognizing this data scarcity, recent methods often lift 2D priors~\citep{kirillov2023segany,tang2024samesh,zhong2024meshsegmenter, liu2023partslip,zhou2023partslip++, yang2024sampart3d,umam2023partdistill,ma20243d} or directly cluster 3D features~\citep{partfield2025} heuristically. Consequently, their partitions are inevitably biased toward data-intrinsic boundaries and are largely governed by low-level geometric cues rather than explicit user intent.  For example, models trained on fixed geometry segmented data rigidly segment an "arm" into sub-parts, failing to align with diverse user intents such as animating the whole arm or editing individual segments. In essence, without multi-granularity data, achieving truly flexible, intent-driven control remains fundamentally out of reach.

Beyond the availability of suitable training data, the mechanism used to formulate and inject user intent is equally critical. While some works attempt to control granularity through global heuristics, such as a scale parameter~\citep{yang2024sampart3d}, a clustering threshold~\citep{tang2024samesh}, or a target part count~\citep{partfield2025}, these mechanisms remain semantically blind, failing to specify which boundaries actually matter. On the other hand, using single-view 2D masks~\citep{li2026segvigenrepurposing3dgenerative} introduces severe view dependence, leaving occluded regions unspecified. Consequently, the reliance on localized or view-dependent signals prevents existing methods from producing coherent, holistic partitions. Closest to our work, recent methods explore control via explicit text schemas~\citep{zhu2026cubepart}. However, they still struggle to offer flexible user-intent control, as these text schemas lack spatial and geometric grounding.

We overcome these two obstacles and unlock flexible, intent-driven control by introducing a \emph{control prompt} as our input signal, which comprises target part names paired with their specific location and shape clauses. This prompt dynamically defines the output label space: varying the prompt directly specifies a coarse, medium, or fine partition, while leaving unspecified regions as background naturally enables selective segmentation. To overcome the lack of multi-granularity training data, we design an automated annotation pipeline powered by a Vision-Language Model (VLM, specifically Gemini). This pipeline analyzes artist-authored components from HY3D-Bench~\citep{hunyuan3d2026hy3dbenchgeneration3dassets} and hierarchically merges them based on semantic and functional relationships to form three granularities, subsequently generating the corresponding control prompts for each level. By applying this pipeline, we annotate 57$K$ objects, yielding a dataset of 150$K$ prompt-target pairs. 

Building upon this large-scale dataset, we introduce \MethodName, a unified framework for text-controlled 3D mesh segmentation.  Our architecture builds upon generative surface attribute models~\citep{Lai_2026_CVPR,li2026segvigenrepurposing3dgenerative} by integrating a part-aware VAE and a text-conditioned Diffusion Transformer (DiT)~\citep{Peebles2022DiT}. Given the input mesh and the control prompt, the DiT generates a latent representation that decodes into a native 3D \emph{part-identity field}. Querying this continuous field at the mesh's face centres yields discrete, binary-encoded part IDs, thereby directly labeling the input without altering its original geometry. Furthermore, to support selective segmentation, we introduce a \emph{drop-to-background} training augmentation that randomly omits parts from the prompt, implicitly assigning their corresponding regions to the background. This diverse training strategy compels the model to follow the semantic prompt rather than being constrained by low-level geometric priors. During inference, a VLM-based planner automatically deduces the location and shape clauses from 2D renderings, unlocking a truly effortless experience where providing simple part names yields precise, intent-driven segmentation.

Our contributions are summarized as follows:
\begin{itemize}[leftmargin=*,nosep]
  \item We introduce a flexible, intent-driven control mechanism via holistic control prompts, and propose an automated VLM-based annotation pipeline to construct a large-scale, multi-granularity dataset with structured part descriptions.
\item We propose \MethodName, formulating text-controlled 3D segmentation as a native 3D part-identity field. Equipped with a randomized background dropping strategy, a single model seamlessly supports multi-granularity partitions and selective segmentation.
\item Extensive experiments demonstrate that \MethodName outperforms evaluated baselines on PartObjaverse-Tiny, achieves state-of-the-art results at each granularity of our held-out \Bench, and significantly improves selective part segmentation.
\end{itemize}

\section{Related Work}
\label{sec:related}

\noindent\textbf{Overcoming 3D data scarcity for part segmentation.}
Existing 3D datasets rarely provide multi-granularity semantic annotations for the same geometry,
so most methods rely on external priors to infer parts. One line of work
distills features from 2D image foundation models~\citep{kirillov2023segany,ravi2024sam,oquab2023dinov2}
into 3D, either by optimizing a per-shape feature field~\citep{yang2024sampart3d} or by predicting
masks on multi-view renderings and lifting them onto the
surface~\citep{tang2024samesh,deng2025geosam2unleashingpowersam2,xu2025sampro3d,Huang2023Segment3D}.
Per-object optimization is costly, and mask fusion is sensitive to view coverage. Open-vocabulary
approaches~\citep{liu2023partslip,zhou2023partslip++,umam2023partdistill,ma20243d,zhong2024meshsegmenter}
leverage vision-language models by querying rendered views with part names and aggregating predictions
back to 3D, but operating in 2D projected space introduces multi-view inconsistencies and fragmented
boundaries on the mesh. A second line learns part features natively in 3D from curated
supervision. PartField~\citep{partfield2025} clusters a learned hierarchy at a chosen level, and
P3-SAM~\citep{ma2025p3sam} predicts point-prompted masks and automatically selects and merges them into a part segmentation. These methods produce
high-quality decompositions, yet because their training data contain only fixed geometric splits,
the resulting partitions are governed by low-level geometric cues rather than user-intended semantic
groupings.

\noindent\textbf{Control signals for part segmentation.}
Given the geometry-biased representations above, various control mechanisms have been proposed, yet all exhibit critical limitations. Global heuristics like scales or target counts~\citep{yang2024sampart3d,tang2024samesh,partfield2025} remain semantically blind, failing to specify which boundaries actually matter. Localized prompts such as clicks or boxes~\citep{zhou2025pointsam,zhu2025partsam,ma2025p3sam} leave semantic grouping implicit, while single-view 2D guidance~\citep{li2026segvigenrepurposing3dgenerative} introduces severe view dependence and leaves occluded regions unspecified. Consequently, these localized or view-dependent signals cannot produce coherent, holistic partitions. Closest to our work, schema-driven approaches~\citep{zhu2026cubepart,pun2026multicube} use text schemas but lack spatial and geometric grounding; furthermore, they synthesize separate geometries rather than segmenting the input surface, and lack multi-granularity training. Unlike these methods,  \MethodName conditions on a holistic text prompt to directly generate a native 3D part-identity field, explicitly trained on multi-granularity annotations to achieve flexible, intent-driven segmentation.

\noindent\textbf{Part-level 3D generation.}
While modern generators excel at producing high-quality holistic shapes~\citep{xiang2024structured,huang2025cupid,lai2025latticedemocratizehighfidelity3d,jia2025ultrashape,zhao2025hunyuan3d,zhang2024clay,zhangb_2023}, recent research has expanded into part-level generation under two main paradigms. Single-stage methods synthesize all parts directly from an image or latent representation, automatically discovering the decomposition~\citep{lin2025partcrafter,tang2024partpacker,chen2025autopartgen,zhang2025bang,chen2024partgen}. Consequently, users lack explicit control over the resulting part structure. Conversely, two-stage methods take an existing shape alongside its decomposition to generate the complete geometry of each individual part~\citep{yan2025xpart,zhao2026structparts,yang2025holopart,yang2025omnipart,sun2026eipartexplodecompletionimplode,fullpart,liao2026megapartsscalingpartaware3d}, meaning their performance is strictly bounded by the input segmentation. Our method provides this segmentation from text, at a user-specified granularity.

\section{Method}
\label{sec:method}

\begin{figure*}[t]
  \centering
  \includegraphics[width=\linewidth]{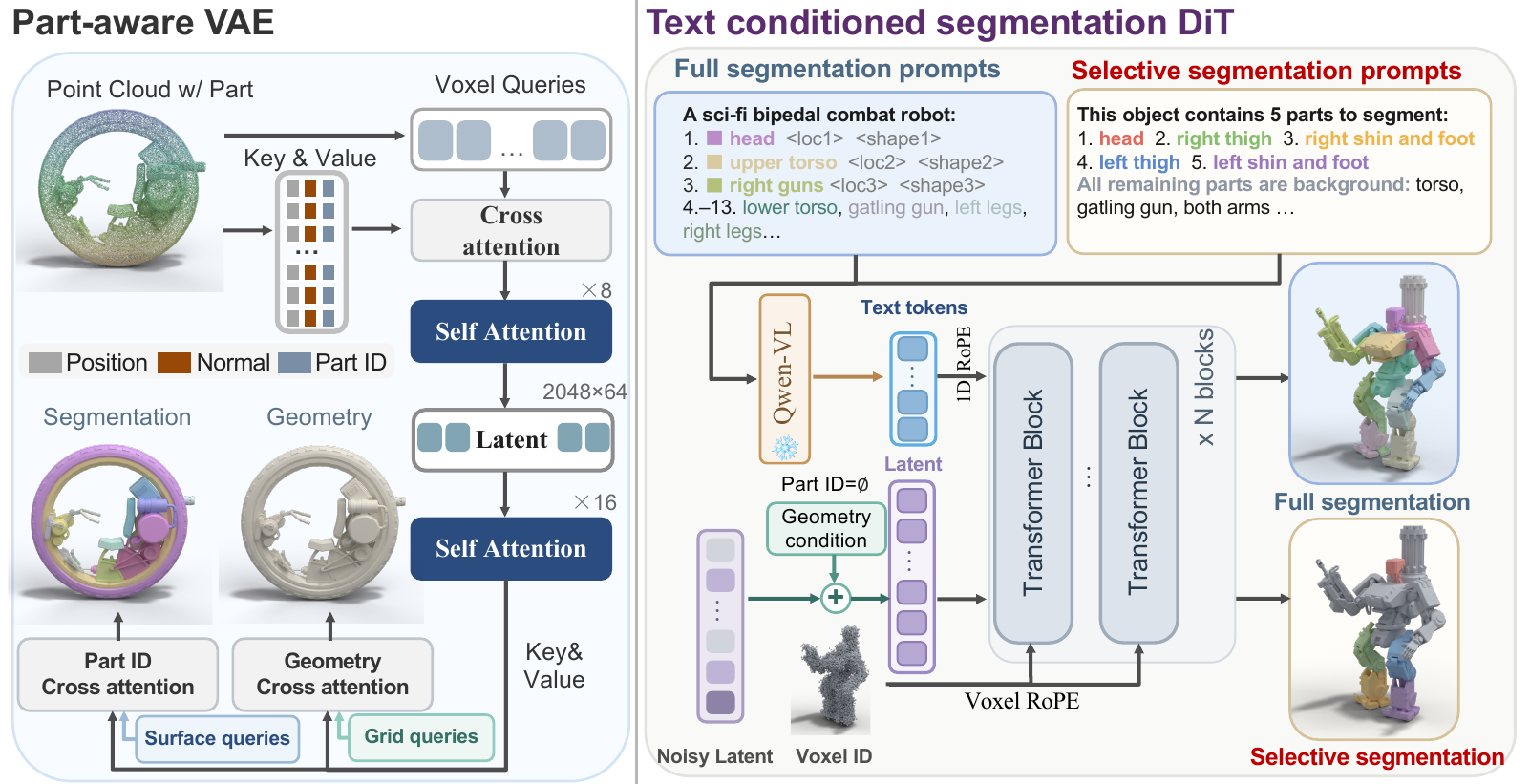}
  \vspace{-2em}
  \caption{\textbf{\MethodName.} \emph{Left:} a part-aware VAE encodes a point cloud carrying
  position, normal, and part ID. Its decoder can be queried at any 3D coordinate, with two cross-attention modules
  for geometry (SDF) and a discrete, binary-encoded part ID. Face-centre queries produce the mesh labels.
  \emph{Right:} a text-conditioned DiT generates the field's latent representation from the
  geometry latent (\textcolor{pipegreen}{green}) and encoded text tokens. A full decomposition (\textcolor{pipeblue}{blue}) and a selective input
  (\textcolor{pipered}{red}) differ only in the control prompts, and in the selective
  case an explicit background item absorbs parts that belong to the background. Each part item is visualized in the color corresponding to the input prompts.
  }
  \vspace{-1.5em}
  \label{fig:method}
\end{figure*}

 As illustrated in \Figref{fig:method}, we propose \MethodName, a unified generative framework for semantics-aware 3D part segmentation driven by a holistic, view-independent text control signal. By simply adjusting the text prompt, users can flexibly control the segmentation granularity and select specific target parts. Building upon native 3D surface fields (\Secref{sec:prelim}), we formulate segmentation as the generation of a continuous 3D part-identity field. Unlike prior methods that predict noisy continuous colors, our field predicts discrete, binary-encoded part IDs that explicitly align with the prompt's label space (\Secref{sec:reformulation}). Our architecture features a part-aware VAE to encode the mesh and its part IDs into a unified latent, alongside a text-conditioned dual-stream DiT that generates these latents from the holistic prompt. Finally, this robust text-to-3D grounding is enabled by training the framework on comprehensive multi-granularity data (\Secref{sec:supervision}).

\subsection{Preliminary}
\label{sec:prelim}

\noindent\textbf{Voxel-anchored latents.}
Following LATTICE~\citep{lai2025latticedemocratizehighfidelity3d}, we encode an input point cloud $\gP$ via cross-attention using anchors at the occupied voxel centres  $\{\voxel_i\}$ of an $\res^3$ grid:
\begin{equation}
\latent = \vaeencoder\big({\voxel_i},\, \gP\big) \in \sR^{n \times d},
\label{eq:voxset}
\end{equation}
where $\vaeencoder$ denotes the encoder. These anchors naturally yield integer grid indices $\rmV$. Following previous work~\citep{zhao2025hunyuan3d,zhao2023michelangelo}, we train a text-conditioned flow-based diffusion model (DiT)~\citep{lipman2024flowmatchingguidecode} on these structured latents. During denoising, the indices $\rmV$ provide spatial guidance by serving as a rotary positional embedding (RoPE)~\citep{su2021roformer} for each token. By explicitly grounding tokens in 3D space, this spatial guidance enables the model to precisely localize features, which is crucial for generating accurate and fine-grained surface attributes.

\noindent\textbf{Dense surface fields.}
Each latent~\citep{lai2025latticedemocratizehighfidelity3d} is anchored at a local voxel, so it carries the information of the surface near that voxel. The decoded latent can therefore produce a continuous field $\sattr$ from $\latent$ at any surface point $\point$,
\begin{equation}
  \sattr(\point) = \decoder\big(\point,\, \latent\big),
  \label{eq:field}
\end{equation}
where $\decoder$ denotes the VAE decoder. The predicted attribute $\sattr$ can represent an SDF for local geometric details, as in LATTICE~\citep{lai2025latticedemocratizehighfidelity3d}, or an SDF and texture color separately, as in NaTex~\citep{Lai_2026_CVPR}. In our method, we reformulate this attribute as the SDF and binary-encoded part IDs, which is detailed in~\Secref{sec:reformulation}.

\subsection{Formulating segmentation as text-controlled part-ID generation}
\label{sec:problem}
\label{sec:reformulation}
Our method takes the input mesh $\mesh$ and a text prompt $\gT$ describing $K$ parts as an ordered list,
optionally followed by a background item describing the remaining parts. The output is a continuous part-ID field; querying it at the face centres of the input mesh with \Eqref{eq:field} yields a per-face label for each of the $F$ faces $\{f_1,\dots,f_F\}$. 

\noindent\textbf{Prompt format.} To provide holistic descriptions of every target part, each item is written as a semantic name with a location and a shape clause, so a full decomposition is conditioned on $\texttt{<sem}_1\texttt{><loc}_1\texttt{><shape}_1\texttt{>}\dots\texttt{<sem}_K\texttt{><loc}_K\texttt{>}\\\texttt{<shape}_K\texttt{>}$. For example, one item might read ``\textit{right arm: upper right side of the torso, a tapering cylindrical limb}.''
A selective prompt keeps the wanted parts in that form and appends the remaining ones as background
items $\texttt{<bg}_1\texttt{><bg}_2\texttt{>}\dots$, each still carrying its own description but
all decoding to the single label $\bg$, as in \Figref{fig:method}.

\noindent\textbf{Binary-encoded part IDs.}
Prior generative segmentation methods represent parts using distinct RGB colors~\citep{li2026segvigenrepurposing3dgenerative}. However, this continuous RGB space $256^3$ often leads to noisy, fragmented predictions within a single part, requiring ad-hoc post-processing such as color clustering to recover part identity. To resolve this, we represent surface attributes using discrete, binary-encoded part IDs $\text{id} \in \{0,1\}^{6}$ to represent the surface attribute that labels the continuous space. Reserving the all-ones label for the background, this 6-bit format supports up to $K= 64-1$ distinct parts, which is sufficient to accommodate the fine-grained structural complexity of most objects. To effectively ground these IDs to user text, we map the $i$-th prompt item to the $i$-th binary ID. Unlike permutation-invariant approaches that lack structural constraints, this locally aligned prompting provides explicit ordering priors, yielding significantly more accurate part assignments.

\noindent\textbf{Part-aware VAE.}
We keep the voxel anchors of \Eqref{eq:voxset} as queries, and the point cloud they attend to
carries a part ID alongside each point's position and normal. An 8-layer self-attention encoder $\vaeencoder$ encodes the shape into a $2048\times64$ latent. The decoder $\decoder$ decodes the latent into fields. As the part field cannot be isolated from geometry information, the latent encodes both geometry and part information. We query the SDF using grid locations that make the latent aware of the location, while querying the six-bit part ID using mesh face centres as queries.
We train the VAE jointly using
\begin{equation}
  \Ls_{\rm{vae}} = \reg_{\rm{kl}} \Ls_{\rm{kl}}
                 + \reg_{\rm{sdf}} \Ls_{\rm{sdf}}
                 + \reg_{\rm{part}} \Ls_{\rm{part}},
  \label{eq:vae}
\end{equation}
where $\Ls_{\rm{kl}}$ is the standard KL divergence regularizing the latent space,  $\Ls_{\rm{sdf}}$ applies a penalty for geometry reconstruction errors, and $\Ls_{\rm{part}}$ uses Binary Cross-Entropy (BCE) to supervise the part IDs.
For VAE training, part indices are randomly shuffled before encoding to prevent order memorization.

\noindent\textbf{Dual-stream DiT.} We employ a dual-stream architecture, inspired by MM-DiT~\citep{lai2025latticedemocratizehighfidelity3d, flux2_2025}, to jointly process the text condition and the 3D latent. First, a frozen Qwen3-VL~\citep{qwen3technicalreport} encoder extracts text tokens \(\tokens\), which are injected with a 1D RoPE~\citep{wu2025qwenimagetechnicalreport}. Simultaneously, to condition on the input geometry, we run the VAE encoder with all part IDs set to zero to produce a part-agnostic shape latent \(\latent_{\rm{s}}\). This geometry condition is added element-wise to the input noisy latent. These latent tokens are then provided with a 3D RoPE~\citep{lai2025latticedemocratizehighfidelity3d,Lai_2026_CVPR} over the voxel indices \(\rmV\) of \Eqref{eq:voxset}. To prevent positional collisions, text indices are strictly offset past the voxel indices. At inference time, the DiT iteratively processes and denoises the latent from noisy $\latentt$ to clean latent $\latentzero$. The clean latent $\latentzero$ is passed to the VAE decoder $\decoder$ to predict the 3D part segmentation with face centres as queries.

\noindent\textbf{Objective.}
We train the model with a flow-matching~\citep{lipman2024flowmatchingguidecode,esser2024scaling}
objective. The target $\latentzero$ is the VAE latent of the part-ID field. Given Gaussian noise
$\noise \sim \gN(\vzero, \mI)$ and a time step $\timestep \sim U[0,1]$, we form the interpolated
state $\latentt = \timestep \latentzero + (1-\timestep)\,\noise$ and supervise the network on the
two conditions above, the text tokens $\tokens$ and the shape latent $\latent_{\rm{s}}$, to predict
the velocity $\latentzero - \noise$,
\begin{equation}
  \Ls_{\rm{fm}} = \E_{\timestep,\, \latentzero,\, \noise}
        \big\lVert \velocity_\parameter\big(\latentt, \timestep, \tokens, \latent_{\rm{s}}\big)
        - (\latentzero - \noise) \big\rVert_2^2 .
  \label{eq:objective}
\end{equation}

\subsection{Data construction}
\label{sec:supervision}
\label{sec:training}

\noindent\textbf{Partitions.}
We take the artist-authored components of
HY3D-Bench~\citep{hunyuan3d2026hy3dbenchgeneration3dassets} as atomic units. Inspired by
CubePart~\citep{zhu2026cubepart}, which uses set-of-mark renderings to group parts, we mark every
component across views and utilize a vision-language model's multi-view reasoning ability for three-level partitions. Duplicate
levels are dropped, so an object used in training yields up to three partitions.

\noindent\textbf{Text annotation.}
For each partition we render every grouped part with its boundary overlay, annotate that part alone
with the clauses of \Secref{sec:reformulation}, then aggregate the results and have a language model
refine them into one prompt. During training we augment it with
\textit{drop-to-background}: a random subset of items is dropped and their faces reassigned to
$\bg$. This is where $\bg$ acquires its meaning, so it is what enables selective segmentation, and
it also strengthens the correspondence between a named item and its label. The detailed pipeline is described in \Appref{sec:appendix_annotation}.

\noindent\textbf{Inference.}
The semantic prompts are either provided directly by the user or automatically generated by a prompt planner. We instantiate this planner using a frozen Qwen3.8~\citep{qwen38} model, which processes multi-view renderings of the input mesh. The planner plays a critical role in selective segmentation scenarios: when a user specifies only a subset of target parts, the planner automatically infers and fills in the remaining background categories, ensuring a complete and valid partition of the entire 3D asset.

\section{Experiments}
\label{sec:experiments}
\newlength{\condimsize}
\setlength{\condimsize}{1.85cm}

\newcommand{\condim}[3]{\includegraphics[trim=#3,clip,width=\condimsize]{figures/3_conditioning_rectangles_20260926/#1/#2.png}}
\newcommand{\condbp}{{\color{black!12}\rule{\condimsize}{\condimsize}}}
\newcommand{\condhdr}[1]{\scriptsize #1}
\newcommand{\condrow}[2]{%
  \condim{#1}{gt}{#2} &
  \condim{#1}{ours_gtlist}{#2} & \condim{#1}{ours_planner}{#2} &
  \condim{#1}{segvigen_gtmap}{#2} & \condim{#1}{segvigen_sam3}{#2} &
  \condim{#1}{p3sam}{#2} & \condim{#1}{partfield}{#2} &
  \condim{#1}{sampart3d}{#2} & \condim{#1}{find3d}{#2} \\}
\newcommand{\condrowbp}{%
  \condbp & \condbp & \condbp & \condbp & \condbp &
  \condbp & \condbp & \condbp & \condbp \\}

\begin{figure*}[!t]
\centering
\setlength{\tabcolsep}{0.5pt}
\renewcommand{\arraystretch}{0.0}

\begin{tabular}{*{5}{c}!{\vrule}*{4}{c}}
\condhdr{GT} & \condhdr{\textbf{Ours-GT}} & \condhdr{\textbf{Ours-Auto}} &
\condhdr{SegviGen-GT} & \condhdr{SegviGen-Auto} & \condhdr{P3-SAM} &
\condhdr{PartField} & \condhdr{SAMPart3D} & \condhdr{Find3D} \\[2pt]
\condrow{d7570691b670496bb670a9fc1d28057c}{70 31 93 86}
\condrow{d34629ae9cf3432188545f36ac1d9cb8}{70 144 93 120}
\condrow{423121778c4b4e849a029b8281c68bda}{70 74 93 86}
\addlinespace[3pt]
\condrow{2084a523a5df4567844524afef5912db}{70 125 93 195}
\end{tabular}
\vspace{-0.5em}
\caption{\textbf{Qualitative comparison on PartObjaverse-Tiny.} Left of the divider, \emph{-GT} denotes ground-truth conditions (our annotated prompt; SegviGen's rendered guidance map), and \emph{-Auto} denotes automatic conditions (our planner-completed names; SegviGen's predicted map). Right of the divider, baselines use their native inputs. As shown, baselines suffer from missing occluded parts (SegviGen), over-segmenting along geometric seams (PartField, P3-SAM, SAMPart3D), or imprecise boundaries (Find3D). \MethodName avoids these issues, effectively following the semantic intent from GT.}
\vspace{-1em}
\label{fig:qualitative}
\end{figure*}

\newlength{\gransmall}      %
\setlength{\gransmall}{1.10cm}
\newlength{\granbig}        %
\setlength{\granbig}{1.65cm}      %
\newlength{\graneq}
\setlength{\graneq}{1.78cm}

\newcommand{\granuidA}{pvxl_8c47c99aee2a4f978dbc24388890997d}
\newcommand{\granuidB}{pvxl_0a3b81274d694e958408d494eaa6b36e}
\newcommand{\granuidC}{pvxl_c02dec85ac0541b8ab6dd56098706b50}
\newcommand{\grantrimA}{60 109 66 167}
\newcommand{\grantrimB}{60 26 66 123}
\newcommand{\grantrimC}{60 139 66 173}
\newcommand{\grantrimD}{124 96 155 80}

\newcommand{\granhdr}[1]{\scriptsize #1}
\newcommand{\gransub}[1]{\tiny #1}
\newcommand{\granlab}[1]{%
  \parbox[b][\graneq][c]{1.2em}{\centering\rotatebox{90}{\scriptsize #1}}}

\newcommand{\granimF}[3]{\includegraphics[trim=#3,clip,width=\granbig]{figures/4_granularity/#1/f_#2.png}}
\newcommand{\granimS}[4]{\includegraphics[trim=#4,clip,width=\gransmall]{figures/4_granularity/#1/#2_#3.png}}
\newcommand{\granbpF}{{\color{black!12}\rule{\granbig}{\granbig}}}
\newcommand{\granbpS}{{\color{black!16}\rule{\gransmall}{\gransmall}}}
\newcommand{\grancell}[3]{%
  \begin{tabular}[b]{@{}c@{}}\granimF{#1}{#2}{#3}\end{tabular}%
  \begin{tabular}[b]{@{}c@{}}\granimS{#1}{c}{#2}{#3}\\[0.6pt]\granimS{#1}{m}{#2}{#3}\end{tabular}}
\newcommand{\grancellbp}{%
  \begin{tabular}[b]{@{}c@{}}\granbpF\end{tabular}%
  \begin{tabular}[b]{@{}c@{}}\granbpS\\[0.6pt]\granbpS\end{tabular}}
\newcommand{\granrowone}[2]{%
  \grancell{#1}{gt}{#2} & \grancell{#1}{ours}{#2} & \grancell{#1}{segvigen}{#2} &
  \grancell{#1}{partfield}{#2} & \grancell{#1}{sampart3d}{#2} & \grancell{#1}{find3d}{#2} \\}
\newcommand{\granrowonebp}{%
  \grancellbp & \grancellbp & \grancellbp &
  \grancellbp & \grancellbp & \grancellbp \\}

\newcommand{\granformatone}{%
\begin{figure*}[!t]
\centering
\setlength{\tabcolsep}{1pt}
\renewcommand{\arraystretch}{0.0}
\begin{tabular}{*{6}{@{\hspace{1.5pt}}c}}
\granhdr{GT} & \granhdr{\textbf{Ours}} & \granhdr{SegviGen} &
  \granhdr{PartField} & \granhdr{SAMPart3D} & \granhdr{Find3D} \\[1pt]
\granrowone{pvxl_8c47c99aee2a4f978dbc24388890997d}{60 109 66 167}
\granrowone{pvxl_0a3b81274d694e958408d494eaa6b36e}{60 26 66 123}
\granrowone{pvxl_c02dec85ac0541b8ab6dd56098706b50}{60 139 66 173}
\addlinespace[3pt]
\granrowone{pvxl_b658acd10ae1432286b3b363fa84afa8}{124 96 155 80}
\end{tabular}
\vspace{-0.5em}
\caption{\textbf{Granularity control on \Bench.} For each method, panels show fine (large), coarse (top right), and medium (bottom right) granularities. While baselines receive their optimal inputs (e.g., GT part counts for PartField), only \Ours (driven by level-specific item lists) successfully adapts its semantic decomposition across all three levels. Conversely, PartField and SAMPart3D are biased toward geometric boundaries over the requested semantic groupings, and Find3D suffers from severe surface fragmentation. Meanwhile, SegviGen struggles to follow the ground truth, since relying on a single reference map fails to resolve occlusions.
}
\vspace{-1.8em}
\label{fig:granularity}
\end{figure*}}

\newcommand{\granimE}[4]{\includegraphics[trim=#4,clip,width=\graneq]{figures/4_granularity/#1/#2_#3.png}}
\newcommand{\granbpE}{{\color{black!12}\rule{\graneq}{\graneq}}}
\newcommand{\grantriple}[3]{%
  \granimE{#1}{c}{#2}{#3} & \granimE{#1}{m}{#2}{#3} & \granimE{#1}{f}{#2}{#3}}
\newcommand{\grantriplebp}{\granbpE & \granbpE & \granbpE}
\newcommand{\granrowtwo}[2]{%
  \granlab{#1} & \grantriple{\granuidA}{#2}{\grantrimA} & \grantriple{\granuidB}{#2}{\grantrimB} &
  \grantriple{\granuidC}{#2}{\grantrimC} \\}
\newcommand{\granrowtwobp}[1]{%
  \granlab{#1} & \grantriplebp & \grantriplebp & \grantriplebp \\}

\newcommand{\granformattwo}{%
\begin{figure*}[!t]
\centering
\setlength{\tabcolsep}{0.5pt}
\renewcommand{\arraystretch}{0.0}
\begin{tabular}{c*{3}{c}@{\hspace{4pt}}*{3}{c}@{\hspace{4pt}}*{3}{c}}
& \gransub{C} & \gransub{M} & \gransub{F} & \gransub{C} & \gransub{M} & \gransub{F} &
  \gransub{C} & \gransub{M} & \gransub{F} \\[2pt]
\granrowtwobp{GT}
\granrowtwobp{\textbf{Ours}}
\granrowtwobp{SegviGen}
\granrowtwobp{PartField}
\granrowtwobp{SAMPart3D}
\granrowtwobp{Find3D}
\end{tabular}
\caption{\textbf{Granularity control on \Bench.} xx
}
\vspace{-5em}
\label{fig:granularity}
\end{figure*}}

\granformatone

\subsection{Experimental setup}
\label{sec:setup}

\noindent\textbf{Training data.}
We run the annotation pipeline of \Secref{sec:supervision} over $57$K objects sampled from
HY3D-Bench~\citep{hunyuan3d2026hy3dbenchgeneration3dassets}, holding the category proportions of
the source library fixed. This yields $151{,}492$ prompt-target pairs.
\Appref{sec:appendix_annotation} provides the annotation details.

\noindent\textbf{Benchmarks.}
We evaluate our method on three main datasets. \textbf{PartObjaverse-Tiny}~\citep{yang2024sampart3d} contains 200 public objects, enabling direct baseline comparisons. Following P3-SAM~\citep{ma2025p3sam}, we use its watertight version. Its original part decomposition is used as given: our pipeline only writes the text description of each part and never edits a part boundary. \textbf{\Bench} is our held-out benchmark of 364 assets proportionally sampled from PartNeXt and PartVerse-XL. Each is annotated at three granularities with strictly no training overlap. \textbf{Selective-190} comprises 190 manually constructed selective cases (ranging from 2 to 18 named parts) across 28 \Bench objects.
We manually inspected and removed objects with defective annotations or source geometries; \Appref{sec:appendix_experiments} details the construction and category distributions of all sets. Finally, we demonstrate that our method generalizes well to generated meshes.

\noindent\textbf{Metrics.}
The baselines do not provide the ordered prompt-to-part-ID correspondence used by \MethodName. We
therefore compare all methods using class-agnostic mIoU~\citep{ma2025p3sam,partfield2025},
averaging each ground-truth part's maximum IoU over predicted regions. For \MethodName, we also
evaluate the same predictions using \emph{-direct} mIoU, which compares output region $k$ with the
ground-truth region for prompt item $k$, without rematching. For selective
segmentation, we report named-part and background IoUs separately to expose foreground leakage.

\noindent\textbf{Baselines.}
We compare against five methods, evaluated in their intended settings (inputs specified in the \emph{Conditioning} column of \Tabref{tab:main_seg}). Find3D~\citep{ma20243d} takes a list of part names, identical to ours. SAMPart3D~\citep{yang2024sampart3d} requires a single scale value, which we calibrate to the benchmark granularities. PartField~\citep{partfield2025} clusters a learned feature field using the ground-truth part count. P3-SAM~\citep{ma2025p3sam} automatically selects and merges predicted masks into a segmentation; we evaluate its native automatic mode without granularity control and exclude it from \Tabref{tab:main_multigran}. SegviGen~\citep{li2026segvigenrepurposing3dgenerative} requires a color-coded guidance map. We evaluate it using both ground-truth maps and maps automatically predicted by prompting SAM3 on a front view with the target part names.
For \Ours, we evaluate a single model using two prompt variations: the annotated item list, and part names enriched with planner-completed location and shape clauses. SegviGen and \Ours are evaluated on both annotated (\emph{-GT}) and automatic (\emph{-Auto}) conditions (\Tabref{tab:main_seg}, \Figref{fig:qualitative}). All text-conditioned methods receive identical English part names. Finally, while CubePart~\citep{zhu2026cubepart} and MultiCube~\citep{pun2026multicube} are the closest schema-conditioned generative methods, MultiCube remains unreleased. As CubePart generates separate per-part meshes rather than per-face labels, direct quantitative comparison is infeasible; we thus provide qualitative comparisons in \Appref{sec:appendix_cubepart}. Detailed protocols are in \Appref{sec:appendix_experiments}.

\begin{table}[t]
  \centering
    \caption{\textbf{Part segmentation on PartObjaverse-Tiny.} \emph{Conditioning} denotes the input signals. \emph{-GT} uses ground-truth annotations, while \emph{-Auto} relies on automated inputs (e.g., predicted color maps or planner-completed lists). \emph{-direct} evaluates our predictions without region-to-GT rematching, and still outperforms all baselines.}
    \vspace{-0.5em}
    \label{tab:main_seg}
    \resizebox{0.8\linewidth}{!}{%
    \begin{tabular}{llc}
      \toprule
      Method & Conditioning & mIoU $\uparrow$ \\
      \midrule
      Find3D~\citep{ma20243d}                                & part names          & 0.282 \\
      SAMPart3D~\citep{yang2024sampart3d}                     & one scale value      & 0.483 \\
      PartField~\citep{partfield2025}                        & target part count & 0.517 \\
      P3-SAM~\citep{ma2025p3sam}                             & none     & 0.560 \\
      \midrule
      SegviGen-GT~\citep{li2026segvigenrepurposing3dgenerative} & part color map & \sbest 0.594 \\
      SegviGen-Auto                                          & part color map, predicted & 0.413 \\
      \midrule
      \Ours-GT                                               & full item list & \best 0.691 \\
      \Ours-Auto                                             & part names only       & \sbest 0.681 \\
      \Ours-direct IoU                                       & full item list & 0.614 \\
      \bottomrule
    \end{tabular}%
    }
\end{table}

\begin{table}[t]
  \centering
    \caption{\textbf{Multi-granularity control.} Methods are evaluated across three granularities on \Bench using their optimal inputs. \Ours significantly outperforms all baselines at every level, and even our \emph{-direct} protocol (\Tabref{tab:main_seg}) exceeds all baselines on average. Although PartField performs reasonably well at the fine level by merely fitting the geometry, all baselines struggle at the coarse and medium levels where semantic understanding is required.}
    \label{tab:main_multigran}
    \renewcommand{\arraystretch}{0.9}%
    \resizebox{0.9\linewidth}{!}{%
    \begin{tabular}{lcccc}
      \toprule
      Method & Coarse $\uparrow$ & Medium $\uparrow$ & Fine $\uparrow$ & Avg. $\uparrow$ \\
      \midrule
      Find3D~\citep{ma20243d}                                & 0.364 & 0.154 & 0.086 & 0.201 \\
      SAMPart3D~\citep{yang2024sampart3d}                     & 0.473 & 0.359 & 0.376 & 0.402 \\
      SegviGen~\citep{li2026segvigenrepurposing3dgenerative} & 0.592 & 0.466 & 0.409 & 0.489 \\
      PartField~\citep{partfield2025}                        & 0.508 & 0.555 & \sbest 0.499 & 0.521 \\
      \midrule
      \Ours                                                  & \best 0.763 & \best 0.635 & \best 0.511 & \best 0.637 \\
      \Ours-direct IoU                                       & 0.751 & 0.615 & 0.467 & 0.611 \\
      \bottomrule
    \end{tabular}%
    }
  \vspace{-1.5em}
\end{table}

\noindent\textbf{Implementation.}
Our implementation builds on the open-source Hunyuan3D~2.1~\citep{hunyuan3d2025hunyuan3d21}. Specifically, we adopt the MM-DiT conditioning, part-aware VAE, and rotary position encoding strategies from LATTICE~\citep{lai2025latticedemocratizehighfidelity3d} and NaTex~\citep{Lai_2026_CVPR}. Text conditioning is processed using a frozen Qwen3-VL-4B-Instruct~\citep{qwen3technicalreport} encoder. During inference, we employ the flow-matching Euler sampler~\citep{esser2024scaling} with 30 steps and a guidance scale of 6.0. Detailed architectural configurations and the training schedule are provided in \Appref{sec:appendix_implementation}.

\subsection{Main results}
\label{sec:main_results}

\noindent\textbf{Evaluation on PartObjaverse-Tiny.}
As shown in \Tabref{tab:main_seg} and \Figref{fig:qualitative}, \Ours achieves the highest accuracy. Lacking semantic control, PartField, P3-SAM, and SAMPart3D rely strictly on geometric boundary priors. Find3D uses text but lacks clean global assignments, while SegviGen's single-view guidance is vulnerable to occlusions. In contrast, our global text-driven approach inherently resolves these occlusions. Furthermore, we evaluate an automated setting (\Ours-Auto) that requires minimal user effort. Users simply provide part names, and our planner automatically completes the detailed location and shape clauses based on renderings. This automation incurs a negligible performance drop of just 0.010 mIoU compared to the fully annotated ground truth (\emph{-GT}). In contrast, SegviGen-Auto suffers a severe 0.181 mIoU drop. Crucially, while SegviGen requires a 2D guidance map that is either tedious to paint or suboptimal to automate, our method only demands a simple text list, ensuring superior usability and robustness. Notably, even under a strict no-rematching protocol, our \emph{-direct} variant still outperforms all baselines (\Tabref{tab:main_seg}).

\noindent\textbf{Granularity control.} 
\Tabref{tab:main_multigran} evaluates multi-granularity control on \Bench, where \Ours outperforms all baselines at every level by simply updating the input text. In contrast, baselines are either impractical (SegviGen requires a new 2D map per level) or semantically blind (PartField specifies only region counts, not identities). Consequently, they remain rigidly tied to geometric boundaries and struggle with semantic abstraction. This is starkly evident in coarse functional grouping, where \Ours surpasses PartField by 0.255 mIoU. As \Figref{fig:granularity} shows, modifying our text prompt seamlessly shifts partitions from broad assemblies to fine-grained accessories. %

\newlength{\selimsize}
\setlength{\selimsize}{1.75cm}

\newcommand{\selim}[3]{\includegraphics[trim=#3,clip,width=\selimsize]{figures/5_selective/#1/#2.png}}
\newcommand{\selbp}{{\color{black!12}\rule{\selimsize}{\selimsize}}}
\newcommand{\selhdr}[1]{\small #1}
\newcommand{\sellab}[2]{%
  \parbox[b][#2][c]{1.2em}{\centering\rotatebox{90}{\small #1}}}
\newcommand{\selrow}[4]{%
  \sellab{#1}{#4} & \selim{#2}{clicks}{#3} & \selim{#2}{ours_gtdesc}{#3} &
  \selim{#2}{ours_planner}{#3} & \selim{#2}{segvigen}{#3} & \selim{#2}{p3sam}{#3} &
  \selim{#2}{find3d}{#3} \\}
\newcommand{\selrowbp}[1]{%
  \sellab{#1}{\selimsize} & \selbp & \selbp & \selbp & \selbp & \selbp & \selbp \\}

\newlength{\ablimsize}
\setlength{\ablimsize}{1.625cm}

\newcommand{\ablim}[3]{\includegraphics[trim=#3,clip,width=\ablimsize]{figures/6_ablation/#1/#2.png}}
\newcommand{\ablbp}{{\color{black!12}\rule{\ablimsize}{\ablimsize}}}
\newcommand{\ablhdr}[1]{\small #1}
\newcommand{\ablrow}[2]{%
  \ablim{#1}{ours}{#2} & \ablim{#1}{wodrop}{#2} & \ablim{#1}{rgb}{#2} \\}
\newcommand{\ablrowbp}{\ablbp & \ablbp & \ablbp \\}

\begin{figure*}[t]
  \centering
  \begin{minipage}[t]{0.655\linewidth}
    \centering
    \setlength{\tabcolsep}{0.5pt}
    \renewcommand{\arraystretch}{0.0}
    \begin{tabular}{c*{6}{c}}
    & \selhdr{Input} & \selhdr{\textbf{Ours}} & \selhdr{\textbf{Ours}} &
      \selhdr{SegviGen} & \selhdr{P3-SAM} & \selhdr{Find3D} \\[-1pt]
    & \selhdr{clicks} & \selhdr{GT desc.} & \selhdr{planner} &
      \selhdr{clicks} & \selhdr{clicks} & \selhdr{names} \\[2pt]
    \selrow{$K{=}4$}{0b1003964aa6_1a2439aa946f4004}{44 48 61 122}{1.525cm}
    \selrow{$K{=}8$}{0146f4c8c5cf_1773e75cc88c5f70}{124 62 89 76}{2.0875cm}
    \end{tabular}
    \vspace{-0.5em}
    \captionof{figure}{\textbf{Selective part segmentation.} The background parts become one brown region. Our two columns denote the same model, evaluated with either the ground-truth prompts or the planner-written descriptions. A click indicates the part the user wants to segment, which may lead to a region with ambiguous boundaries, as seen in P3-SAM and SegviGen.}
    \label{fig:selective}
  \end{minipage}\hfill
  \begin{minipage}[t]{0.325\linewidth}
    \centering
    \setlength{\tabcolsep}{0.5pt}
    \renewcommand{\arraystretch}{0.0}
    \begin{tabular}{*{3}{c}}
    \ablhdr{\textbf{Ours}} & \ablhdr{w/o drop} & \ablhdr{RGB} \\[2pt]
    \ablrow{1d78293d14124949ac3c18d1ffeb4490}{0 70 161 74}
    \ablrow{3aac7731452ff98155423f097fd5931f1802137c495ff185c53bc4637df62061}{0 39 191 74}
    \end{tabular}
    \vspace{-0.5em}
    \captionof{figure}{\textbf{Ablation study of two design choices.} The RGB palette
    fragments parts that part IDs keep whole (rectangles); w/o drop-to-background training reduces accuracy.}
    \label{fig:ablation}
  \end{minipage}
  \vspace{-0.5em}
\end{figure*}

\begin{table}[t]
  \centering
  \vspace{0.8em}
  \vspace{-1em}
    \caption{\textbf{Selective part segmentation.} Query IoU averages over the named parts, and BG IoU covers the background region. \MethodName outperforms all click-driven baselines, even when relying on planner-generated prompts from names only. While baseline methods using solely part names (Find3D) fail to extract user-intended parts, our control prompts successfully achieve accurate selective segmentation.
    }
    \vspace{-1em}
    \label{tab:selective}
    \resizebox{0.9\linewidth}{!}{%
    \begin{tabular}{llcc}
      \toprule
      Method & Input & Query IoU $\uparrow$ & BG IoU $\uparrow$ \\
      \midrule
      Find3D~\citep{ma20243d}                                & part names & 0.090 & 0.745 \\
      P3-SAM~\citep{ma2025p3sam}                             & GT clicks & 0.230 & 0.413 \\
      SegviGen~\citep{li2026segvigenrepurposing3dgenerative} & GT clicks & 0.456 & 0.764 \\
      \midrule
      \Ours (planner)  & names only        & \sbest 0.615 & \sbest 0.895 \\
      \Ours (GT desc.) & full descriptions & \best 0.657 & \best 0.933 \\
      \bottomrule
    \end{tabular}%
    }
    \vspace{-1em}
\end{table}

\begin{table}[t]
  \centering
    \caption{\textbf{Ablation.} \textit{Full} averages the three granularities on a human category subset from \Bench; \textit{Selective} is the accuracy on Selective-190. Using an RGB palette instead of part IDs fragments cohesive regions, resulting in lower accuracy across both tasks. Furthermore, drop-to-background training is vital; it facilitates semantics-aware understanding and strictly enables selective segmentation.
    }
    \vspace{0em}
    \label{tab:ablation}
    \renewcommand{\arraystretch}{0.8}%
    \setlength{\tabcolsep}{0.6em}%
    \resizebox{0.66\linewidth}{!}{%
    \begin{tabular}{lcc}
      \toprule
      & Full $\uparrow$ & Selective $\uparrow$ \\
      \midrule
      \Ours & \best 0.845 & \best 0.657 \\
      w/o drop-to-background & 0.818 & \xmark \\
      \midrule
      RGB, not Part IDs & 0.776 & 0.646 \\
      \bottomrule
    \end{tabular}%
    }
  \vspace{-1.5em}
\end{table}

\newlength{\aigcimsize}
\setlength{\aigcimsize}{2.15cm}
\newcommand{\aigcim}[3]{\includegraphics[trim=#3,clip,width=\aigcimsize]{figures/7_aigc/#1/#2.png}}
\newcommand{\aigcbp}{{\color{black!12}\rule{\aigcimsize}{\aigcimsize}}}
\newcommand{\aigchdr}[1]{\small #1}
\newcommand{\aigcsub}[1]{\scriptsize #1}
\newcommand{\aigctriple}[2]{\aigcim{#1}{c}{#2} & \aigcim{#1}{m}{#2} & \aigcim{#1}{f}{#2}}
\newcommand{\aigctriplebp}{\aigcbp & \aigcbp & \aigcbp}
\newcommand{\aigcrow}[4]{\aigctriple{#1}{#2} & \aigctriple{#3}{#4} \\}
\newcommand{\aigcrowbp}{\aigctriplebp & \aigctriplebp \\}

\newlength{\failimsize}
\setlength{\failimsize}{2.0625cm}
\newcommand{\failim}[3]{\includegraphics[trim=#3,clip,height=\failimsize]{figures/8_failure/#1/#2.png}}
\newcommand{\failbp}{{\color{black!12}\rule{\failimsize}{\failimsize}}}
\newcommand{\failhdr}[1]{\small #1}
\newcommand{\failrow}[2]{\failim{#1}{gt}{#2} & \failim{#1}{ours}{#2} \\}
\newcommand{\failrowbp}{\failbp & \failbp \\}

\begin{figure*}[t]
  \centering
  \begin{minipage}[t]{0.78\linewidth}
    \centering
    \setlength{\tabcolsep}{0.5pt}
    \renewcommand{\arraystretch}{0.0}
    \begin{tabular}{*{3}{c}@{\hspace{0.25cm}}*{3}{c}}
    \aigcsub{Coarse} & \aigcsub{Medium} & \aigcsub{Fine} &
      \aigcsub{Coarse} & \aigcsub{Medium} & \aigcsub{Fine} \\[2pt]
    \aigcrow{2e686eb0-1036-4fe5-a372-49a070058871}{61 72 66 96}%
          {4d1750c2-db2b-427b-9d42-80fa742ab6f5}{61 72 66 96}
    \aigcrow{08671bba-5610-4b12-b15f-a79a30d5a9f8}{86 79 87 93}%
          {2357c8e5-e5ed-4b8e-aa50-33a1157978cb}{3 116 230 114}
    \end{tabular}
    \vspace{-0.5em}
    \captionof{figure}{\textbf{Generalization to generated assets.} Four Hunyuan3D~3.1 meshes. Although these AI-generated meshes lack well-defined geometric boundaries, our method generalizes robustly by following semantic instructions rather than relying on rigid geometric priors.
    }
    \label{fig:aigc}
  \end{minipage}\hfill
  \begin{minipage}[t]{0.20\linewidth}
    \centering
    \setlength{\tabcolsep}{0.5pt}
    \renewcommand{\arraystretch}{0.0}
    \begin{tabular}{*{2}{c}}
    \failhdr{GT} & \failhdr{\textbf{Ours}} \\[2pt]
    \failrow{repeated}{175 94 166 148}
    \failrow{vague}{192 82 191 84}
    \end{tabular}
    \vspace{-0.5em}
    \captionof{figure}{\textbf{Failure cases} with vague descriptions (location and shape).}
    \label{fig:failure}
  \end{minipage}
  \vspace{-1.5em}
\end{figure*}

\noindent\textbf{Selective part segmentation.} \Tabref{tab:selective} evaluates selective segmentation, where specific target parts are isolated and the rest merged into a single background region (\Figref{fig:selective}). Because click-driven baselines cannot handle this semantic grouping, we supply them with ground-truth points inside each target part. Nevertheless, clicks only provide localization without semantic context. As a result, these baselines fail to capture the user's intent, struggling to separate the targets from the background cleanly. Find3D accepts text but similarly suffers from noisy boundary delineation. Our method uniquely satisfies user intent by assigning unselected geometry to a unified background region. Finally, our planner's automated name expansion recovers most of the accuracy of ground-truth descriptions.

\noindent\textbf{Ablations.} \Tabref{tab:ablation} validates our control prompt designs. First, drop-to-background training is essential for selective segmentation and improves full accuracy (0.818 to 0.845); explicitly prompting specific parts to the background label strengthens prompt-output alignment. Second, Part IDs strictly outperform an RGB palette. As \Figref{fig:ablation} shows, continuous RGB spaces cause decoding ambiguity and surface fragmentation, whereas discrete Part IDs ensure clean, cohesive boundaries. We further ablate multi-granularity training and prompt--label order binding in
\Appref{sec:appendix_training_diagnostics}. Moreover, a fixed-geometry, fixed-$K$ prompt ablation verifies that the model uses the requested semantics rather than only the number of regions (\Appref{sec:appendix_semantic_ablation}).

\noindent\textbf{Generalization to generated assets.} To demonstrate generalization beyond artist-made assets, we evaluate on AI-generated meshes (Hunyuan3D 3.1), which present challenging topologies and boundary qualities (\Figref{fig:aigc}). Lacking ground-truth annotations, our automated planner autonomously constructs coarse, medium, and fine prompts relying solely on renderings and a target granularity. Granularity control transfers seamlessly to these generated assets: coarse prompts successfully separate broad functional groups, while fine prompts produce atomic decomposition.

\section{Conclusion}
\label{sec:conclusion}

We advance the user-centric principle for 3D part segmentation: \textbf{what you ask for is what you get}. We present \MethodName, a unified text-controlled framework that overcomes two core obstacles: the lack of multi-granularity data and the absence of flexible, intent-driven control. We introduce a holistic prompt to dynamically define the target label space. A large-scale, VLM-annotated dataset supports this design, enabling text-guided segmentation. Generating a native 3D part-identity field conditioned on these prompts, combined with randomized background dropping, ensures the model effectively follows semantic instructions rather than rigid geometric priors. Extensive experiments demonstrate \MethodName's significant superiority over strong baselines in both multi-granularity and selective segmentation tasks.

\noindent\textbf{Limitations and future work.}
\Figref{fig:failure} illustrates limitations in resolving part boundaries and distinguishing
similar parts. Location clauses may not uniquely identify repeated or symmetric components,
merging two of a ship's five sails, and overlapping descriptions could leave the boundary between
adjacent parts ambiguous, as between a knight's legs and leg armour. Even when the
model recovers the intended regions, it could assign their identifiers incorrectly,  as reflected in the -direct  results in \Tabref{tab:main_seg} and \Tabref{tab:main_multigran}.
A promising direction is an agentic workflow in which a
language model and the user iteratively refine the prompt after inspecting the returned partition.

{
    \small
    \bibliographystyle{ieeenat_fullname}
    \bibliography{main}
}

\clearpage
\onecolumn
\begin{center}
  {\Large\textbf{\thetitle}}\\[0.5em]
  {\Large Supplementary Material}\\[1.0em]
\end{center}
\appendix
\section{Additional Details}
\label{sec:appendix}

\subsection{Annotation pipeline}
\label{sec:appendix_annotation}
This appendix describes the pipeline introduced in \Secref{sec:supervision}, in the same three
stages: the partitions, the text annotation of each part, and the planner that replaces the
annotation at inference.

\paragraph{Stage 1: partitions.}
We take the artist-authored components of each asset as atomic units, since an artist has already
split the mesh where a part boundary belongs, and the stage decides only which components belong
together. Every component is marked across the sixteen views
(\Figref{fig:annotation_som}) and one vision-language call returns all three levels, with the
component IDs as its vocabulary:

\begin{promptbox}
The 3D asset has {n} parts (numbered 0 to {n-1}).
Produce THREE levels of semantic clustering in one pass:

COARSE -- top-level functional assemblies: each group is a complete,
  independently meaningful sub-system (e.g. "wheels", "left arm").
  Reference range: ~3-7 groups; let the object's structure decide.
MIDDLE -- natural subdivisions within each coarse group. Separate
  left/right symmetric pairs and functionally distinct sub-parts.
  If a coarse group is a single atomic part, its medium entry may be
  identical. Reference range: ~7-15 groups.
FINE   -- every semantically distinct part, with precise names.
  Reference range: ~12-20 groups.

Every part ID must appear in exactly one group at each level.
Return ONLY this JSON:
{
  "object_type": "brief description of the asset",
  "coarse": [{"name": "name", "part_ids": [0, 1]}, ...],
  "medium": [{"name": "name", "part_ids": [0]},   ...],
  "fine":   [{"name": "name", "part_ids": [0]},   ...]
}
\end{promptbox}

The three levels are requested independently, so a coarse group is the functional grouping an
artist would ask for and need not be a union of the fine groups. A level identical to another level
is dropped, so an object yields up to three partitions. Each surviving level is one segmentation
target: the \texttt{part\_ids} of its $k$-th group become label $k$ on the faces of those
components.

\newlength{\annimsize}
\setlength{\annimsize}{5.25cm}
\newcommand{\annim}[1]{\includegraphics[width=\annimsize]{figures/9_annotation/#1.png}}

\begin{figure}[h]
  \centering
  \setlength{\tabcolsep}{2pt}
  \renewcommand{\arraystretch}{0.0}
  \begin{tabular}{cc}
    \annim{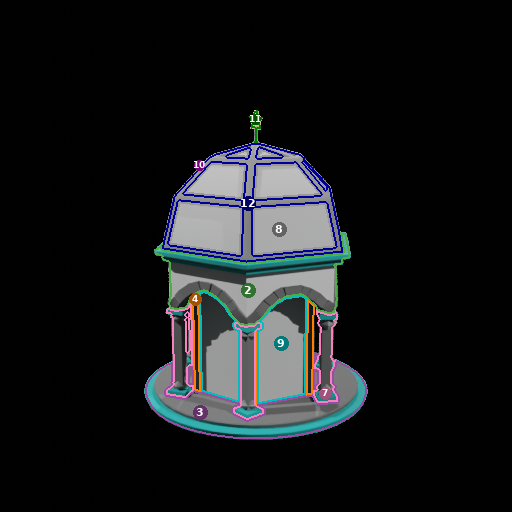} & \annim{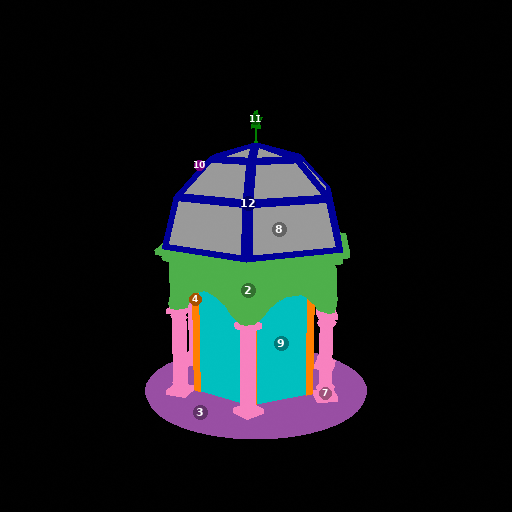} \\[2pt]
    {\small Textured, with marks} & {\small Part-colored, with marks} \\
  \end{tabular}
  \caption{\textbf{What the grouping call sees, per view.} Each component carries a numbered mark in
  both renders, so the returned groups are lists of IDs. The two renders disagree about where a
  boundary is only where a texture hides one, which is why both are given. Sixteen such pairs cover
  the object, and one call returns all three levels.}
  \label{fig:annotation_som}
\end{figure}

\paragraph{Stage 2: text annotation.}
This stage turns one partition into one prompt, one group per call.
\Figref{fig:stage2_call} shows a complete request. Each group is rendered from four viewpoints using three
rendering modes: its boundary outlined on the whole object, which places it in context; the group on its
own, which shows its shape without occlusion; and the full scene, which is the object it belongs
to. Four viewpoints cover what a single view hides. The object-level caption, produced once per
asset, is supplied as read-only context.

The call returns one JSON object per group, with the group ID echoed back so the answer and its
geometry target stay paired:

\begin{promptbox}
{
  "SourceGroupId": "Group_01",         // echoed, must match the request
  "PartCategory":  "arm",              // high-level category of the group
  "PartName":      "right arm",        // unique within this object
  "PartPhrase":    "right arm with pauldron and gauntlet",
  "SpatialRelation": "upper right side of the torso, extending downward",
  "Shape": "a tapering cylindrical limb with a rounded shoulder plate"
}
\end{promptbox}

The category and name must cover every recognisable component of the group instead of its most
prominent one: where an upper arm, forearm, gauntlet and hand together form one limb, the category
is ``arm''. A second, text-only call then aggregates the per-group answers into the prompt the
model trains on, mapping \texttt{PartName} to the item name, \texttt{SpatialRelation} to the
location clause and \texttt{Shape} to the shape clause, in group-ID order and with the object
caption as the preamble. This aggregation call receives only the per-group responses:

\begin{promptbox}
Below are per-part descriptions of one object, with their ids.
Write a single prompt: one sentence naming the object and its part count,
then the items in the given id order, each as "name: location, geometry".
Keep the given ids and their order. Introduce no detail that is not in
the input.
\end{promptbox}

\newlength{\stagetwow}
\setlength{\stagetwow}{2.0875cm}
\newcommand{\stagetwoim}[2]{\includegraphics[width=\stagetwow]{figures/11_stage2/#1_v#2.png}}
\newcommand{\stagetworow}[2]{%
  \rotatebox{90}{\scriptsize #2} &
  \stagetwoim{#1}{02} & \stagetwoim{#1}{13} & \stagetwoim{#1}{14} & \stagetwoim{#1}{15} \\}

\begin{figure}[h]
  \centering
  \begin{minipage}[c]{0.58\linewidth}
    \centering
    \setlength{\tabcolsep}{1.5pt}
    \renewcommand{\arraystretch}{0.0}
    \begin{tabular}{c*{4}{c}}
      \stagetworow{overlay}{overlay}
      \stagetworow{solo}{solo}
      \stagetworow{fullscene}{full scene}
    \end{tabular}
  \end{minipage}\hfill
  \begin{minipage}[c]{0.40\linewidth}
\begin{promptbox}
Object: a walker mech. It has 5 parts.
Annotate ONLY group Group_01.

For four viewpoints you are given:
  overlay    - the group outlined
               on the whole object
  solo       - the group on its own
  full scene - the object, for
               reference

Return JSON:
{
  "SourceGroupId": "Group_01",
  "PartCategory": "...",
  "PartName": "...",
  "PartPhrase": "...",
  "SpatialRelation": "...",
  "Shape": "..."
}
Cover every recognisable component
of the group, not only its most
prominent one. Echo SourceGroupId.
\end{promptbox}
  \end{minipage}
  \caption{\textbf{One annotation call.} The images and the prompt a single Stage~2 request
  receives, for the \texttt{left\_arm} group of a five-group coarse partition. The overlay places
  the group in the whole object, the solo render shows its shape without occlusion, and the full
  scene is the object it belongs to; four viewpoints of each cover the parts a single view hides.
  The call returns one JSON object, and the aggregation step maps \texttt{PartName},
  \texttt{SpatialRelation} and \texttt{Shape} to the item's three clauses.}
  \label{fig:stage2_call}
\end{figure}

\paragraph{Consistent location clauses.}
A clause that says ``left'' is only meaningful if the object has a front to be left of, and if the
annotator agrees with the reader about where that front is. Turntable renderings make this easy to
get wrong: the same part is on the left in one view and on the right in another. We therefore
estimate, once per object, whether it has a well-defined front, using
Orient-Anything~\citep{wang2024orient} over the same multi-view rig the planner sees at inference.
That decision fixes which vocabulary the \texttt{SpatialRelation} field may use:

\begin{promptbox}
directional    : You may use global directions -- left, right, front, back --
                 which are fixed for this object across all views.
non-directional: Use only relative terms: above, below, between, or adjacent
                 to a named part.
\end{promptbox}

A directional object thus keeps one handedness across all sixteen views, and a symmetric one, a
vase for instance, is described by relations among its own parts.

\subsection{Additional ablation studies}
\label{sec:appendix_training_diagnostics}

\paragraph{Multi-granularity training.}
To evaluate the importance of multi-granularity training for semantics-aware control, we compare our approach against a baseline trained solely on fine-level partitions (\Tabref{tab:human_multigran_training}). As observed, training with only a single segmentation pattern per geometry limits the model's flexibility. While it maintains adequate fine-level understanding, it performs poorly on coarse and medium granularities. By contrast, multi-granularity training significantly improves performance across all semantic levels—raising the average mIoU from 0.708 to 0.845. This clearly indicates that multi-granularity data is essential for empowering the model with robust, intent-driven segmentation capabilities.

\begin{table}[H]
  \centering
  \caption{\textbf{Multi-granularity training.}}
  \label{tab:human_multigran_training}
  \small
  \setlength{\tabcolsep}{0.5em}
  \begin{tabular}{lcccc}
    \toprule
    Training & Coarse $\uparrow$ & Medium $\uparrow$ & Fine $\uparrow$ & Avg. $\uparrow$ \\
    \midrule
    w/o multi-granularity training & 0.606 & 0.753 & 0.766 & 0.708 \\
    w/ multi-granularity training  & 0.891 & 0.852 & 0.792 & 0.845 \\
    \bottomrule
  \end{tabular}
\end{table}

\paragraph{Prompt--label order binding.}

\Tabref{tab:human_label_order} compares unordered and ordered training on the same multi-granularity data. Unordered training randomly assigns target IDs and shuffles prompt lists, capturing only global prompt-partition correlations. In contrast, ordered training aligns target IDs with their corresponding prompt positions, modeling both global partition structures and local part-to-label correlations. The results clearly show that unordered training fails to establish direct semantic-to-channel mapping, resulting in a near-zero Direct mIoU (0.011). Ordered training not only restores direct prediction capabilities (Direct mIoU of 0.804) but also achieves a higher Matched mIoU (0.845 vs. 0.767). This indicates that explicitly enforcing local semantic-latent binding intrinsically improves the model's overall understanding ability.

\begin{table}[H]
  \centering
  \caption{\textbf{Ordered and unordered training.}}
  \label{tab:human_label_order}
  \small
  \setlength{\tabcolsep}{0.8em}
  \begin{tabular}{lcc}
    \toprule
    Training & Matched mIoU $\uparrow$ & Direct mIoU $\uparrow$ \\
    \midrule
    Unordered & 0.767 & 0.011 \\
    Ordered   & 0.845 & 0.804 \\
    \bottomrule
  \end{tabular}
\end{table}

\subsection{Implementation details}
\label{sec:appendix_implementation}
\paragraph{Architecture.}
The denoiser is an MM-DiT with 12 double-stream and 24 single-stream blocks, hidden size 1536 and
12 heads. Text is encoded by a frozen Qwen3-VL-4B-Instruct~\citep{qwen3technicalreport}, whose last
hidden state is projected from 2560 to 1536 by a trainable RMSNorm and linear layer under a
1024-token budget. The part-ID field is encoded by the part-aware VAE of
\Secref{sec:reformulation} at a resolution of $128$, and geometry and prompt tokens are indexed in
one sequence by the rotary encoding of \Secref{sec:reformulation}. Text is dropped for $10\%$ of
training samples, which provides the unconditional branch for classifier-free guidance.

\paragraph{Training.}
AdamW, base learning rate $10^{-4}$ with $\beta=(0.9,0.99)$, $\epsilon=10^{-6}$ and weight decay
$10^{-2}$, a cosine schedule with $1{,}000$ warmup steps, gradient clipping at $0.5$, in fp32. The
text projection trains at $5\times10^{-5}$ while the denoiser trains at $10^{-4}$. We train for
$125$K steps. At inference we use the flow-matching Euler sampler with $30$ steps and guidance
scale $6.0$.

\paragraph{Prompt perturbations.}
\Secref{sec:training} introduces drop-to-background; a second perturbation is applied alongside it,
and both act on the prompt.
\emph{Drop-to-background} is applied to every training sample, dropping a fraction of the items
drawn uniformly up to $0.65$ and reassigning their faces to $\bg$.
\emph{Part-list shuffling} permutes the items and relabels the target accordingly, so an index
denotes the item's position in the prompt list.

\paragraph{What the planner sees.}
The planner is Qwen3.8~\citep{qwen38}, held frozen; it is a separate model from
the Qwen3-VL-4B-Instruct that encodes the prompt inside \Ours. Its visual input is the geometry
alone, sixteen turntable renderings at $512^2$ of the bare mesh under a neutral matcap shading
(\Figref{fig:planner_views}).

It is used in two regimes, differing in what the user supplies. When the user states the part names,
as in the \Ours-Auto row of \Tabref{tab:main_seg} and the planner row of \Tabref{tab:selective}, the
planner completes each name into a full item by writing its location and shape clauses, in the field
order the annotation stage of \Appref{sec:appendix_annotation} would have produced, which \Ours
consumes unchanged; the number of parts is then whatever the user listed. For a selective request it
also appends the background items covering the remainder of the surface, which the user does not
enumerate. When no part names are specified, as for the generated assets of \Figref{fig:aigc} where no
ground-truth decomposition exists, the planner receives only the renderings and a target
granularity and decides the inventory itself, so it is not told how many parts to return.

\newlength{\plannerimsize}
\setlength{\plannerimsize}{4.2cm}
\newcommand{\plannerim}[1]{\includegraphics[width=\plannerimsize]{figures/10_planner/#1.png}}

\begin{figure}[h]
  \centering
  \setlength{\tabcolsep}{1.5pt}
  \renewcommand{\arraystretch}{0.0}
  \begin{tabular}{cccc}
    \plannerim{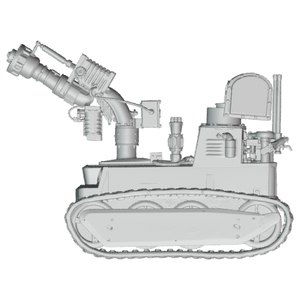} & \plannerim{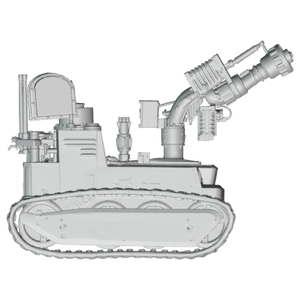} & \plannerim{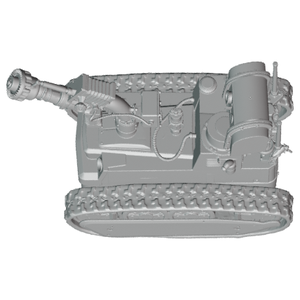} & \plannerim{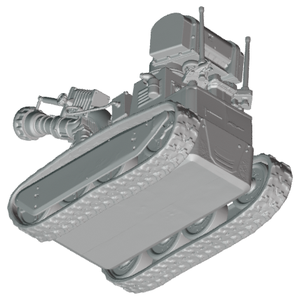} \\
  \end{tabular}
  \caption{\textbf{What the visual planner is given.} Four of the sixteen turntable views, $90^\circ$
  apart, of the bare mesh; these are its only visual input, alongside the user's command. This
  object is also the first row of \Figref{fig:aigc}.}
  \label{fig:planner_views}
\end{figure}

\subsection{Additional experiments}
\label{sec:appendix_experiments}

\paragraph{Benchmark construction.}
\Secref{sec:setup} names the three evaluation sets; this paragraph says how each was built.
For \textbf{PartObjaverse-Tiny} we follow the watertight protocol of P3-SAM~\citep{ma2025p3sam},
which introduces a watertight version of this benchmark and reports baselines on it: every asset is
converted once, each new face taking the label of its nearest original face, and that mesh is the
only geometry any method sees. It is the stricter setting, since the original connected components
correlate with the parts and can be exploited whereas a watertight surface has neither connectivity
nor color. On this benchmark our pipeline supplies text only: the part decomposition released with
PartObjaverse-Tiny is used without modification, and every ground-truth region in
\Tabref{tab:main_seg} and \Tabref{tab:semantic_ablation} is a set of its original parts, so no part
boundary is of our making. What our pipeline adds is the object caption and the name, location and
shape clauses each region is referred to by, which the original release does not provide and which
text-conditioned segmentation requires. \textbf{Selective-190} is built by hand from $28$ \Bench objects: for each case we
choose a subset of the annotated parts as the named items and assign the remainder to background, so
the ground truth is the benchmark's own annotation restricted to that subset rather than a new
labelling. The number of named parts per case runs from two to eighteen, concentrated on the even
values between two and ten. \Bench itself is described below.

\paragraph{Constructing \Bench.}
\Bench is assembled in four stages. We first shortlist candidates from PartNeXt and PartVerse-XL,
scoring each asset on whether its artist-authored components support three distinct levels, and
sample $400$ objects in the category proportions of the source libraries. Sampling targets the fine
part count as well: most objects are asked to land between $20$ and $28$ fine parts, with a capped
tail above that, so the benchmark is not dominated by either trivial or extreme decompositions.
Second, every asset passes the grouping and annotation pipeline of
\Appref{sec:appendix_annotation}, and each of its three levels is checked automatically against a
granularity contract: the three counts must strictly increase, the three partitions must be
pairwise distinct, and levels with excessive singleton
groups, a dominant catch-all group, or micro-fragmented fine partitions are rejected. Because the
contract requires all three levels to survive, every object kept in \Bench carries exactly three
partitions, whereas a training object yields up to three once duplicate levels are dropped
(\Secref{sec:supervision}). No asset
carried a hard error at this stage; the automatic checks instead flagged $58$ of the $400$ as
borderline for a human to read.

Third, and this is what the benchmark rests on, the authors reviewed the annotations by hand ---
every borderline case, and the accepted ones by sampling --- against the rendered partitions rather
than the text alone. We removed $23$ objects outright: $21$ whose source geometry or component
labelling was itself defective, which no annotation could repair, and $2$ more the reviewer judged
mislabelled. A further $13$ were dropped for having more fine parts than the sampling policy allows,
leaving $364$ objects and $1{,}092$ annotated partitions ($152$ from PartNeXt, $212$ from
PartVerse-XL). \Tabref{tab:bench_categories} reports the category distribution; the fine level carries
between $10$ and $28$ parts with a median of $23$. Fourth, the set was frozen and every UID checked
against the training pool.

The filtering and the freeze both predate any model run on this benchmark, so no selection here
could have been informed by which cases our method gets right.

\begin{table}[t]
  \centering
  \caption{\textbf{Category distribution of \Bench} over its $364$ objects, following the
  proportions of the source libraries.}
  \label{tab:bench_categories}
  \small
  \begin{tabular}{lr@{\hspace{2.5em}}lr}
    \toprule
    Category & \# & Category & \# \\
    \midrule
    Human                & 91 & Building         & 16 \\
    Weapon and tool      & 66 & Furniture and home & 15 \\
    Vehicle              & 58 & Plant            & 10 \\
    Daily container      & 36 & Instrument, other & 5 \\
    Electronics          & 32 & Food             &  5 \\
    Animal               & 30 & \textbf{Total}   & \textbf{364} \\
    \bottomrule
  \end{tabular}
\end{table}

\paragraph{Baseline protocols.}
Each baseline is run with its released weights in the regime it was designed for;
\Secref{sec:setup} says what each is conditioned on and this paragraph says how.
\textbf{SegviGen} releases three checkpoints, one interactive and two for full segmentation with
and without 2D guidance. Both of its \Tabref{tab:main_seg} arms supply a guidance map, so both use
the 2D-guidance checkpoint; \Tabref{tab:selective} uses the \emph{interactive} checkpoint, since a
click on each named part is the closest equivalent it has to a selective input. Its predicted-map
arm queries SAM~3 on a single front view with an English translation of the annotated part names,
index-locked to the prompt. \textbf{SAMPart3D} requires one test-time optimization per object,
$5{,}000$ steps, and emits all of its scales from that single run; the scale-to-granularity
correspondence was fixed on a separate calibration set and held there for both benchmarks.
\textbf{Find3D} resamples the mesh to $5{,}000$ points internally, which bounds the boundary
precision it can reach, and decodes a multi-query panoptic argmax. \textbf{PartField} clusters its
feature field to a part count we supply from the ground truth. \textbf{P3-SAM} is run at its native
part count. Every text-conditioned method, Find3D and both of our rows included, receives the same
English part names.

\paragraph{Background leakage on Selective-190.}
\Tabref{tab:selective} reports Query IoU and BG IoU. A third quantity, the fraction of
ground-truth background area a method predicts as some named part, measures the same failure from
the other side and is what makes the BG IoU column interpretable. \Ours leaks $0.024$ of the
background into its named parts, and Find3D $0.171$. The comparison worth making is against a model trained without
drop-to-background, which leaks $0.976$: it partitions the whole surface as usual and its $\bg$
label carries no meaning, which is the ablation \Tabref{tab:ablation} marks \xmark.

\subsection{Does the model use semantics beyond part count?}
\label{sec:appendix_semantic_ablation}
A potential shortcut is that the model might infer the desired granularity solely from the number of prompt items, 
$K$, while ignoring their actual semantic content. To investigate this, we evaluate on the PartObjaverse-Tiny benchmark (\Secref{sec:setup}), strictly controlling for input geometry, target partition, 
$K$, model checkpoint, sampler, and random seed. Using the \Ours-GT results from \Tabref{tab:main_seg} as our full-prompt baseline, we compare the original control prompt for each object against two alternatives: (1) a list of 
$K$ generic placeholders, and (2) a mismatched prompt sampled from a fixed derangement of genuine annotations with the same 
$K$. For the mismatched prompts, we replace the entire text—including the object caption, part names, location clauses, and shape clauses—while keeping $K$ unchanged. This ensures the prompt retains a natural linguistic structure and annotation format, confirming that any observed effects are not merely artifacts of an unusual placeholder format. For instance, a three-item prompt for an orc figurine is replaced by a genuine three-item sailboat prompt. For pools containing only a single prompt of length $K$, we borrow genuine same-$K$ text from another benchmark partition where possible; for the single global-maximum $K$, we apply a fixed cyclic permutation to its item descriptions. Excluding this specific edge case does not alter our overall conclusions. Finally, we aggregate the evaluation metrics per object and report the average across all objects.

\begin{table}[h]
  \centering
  \caption{\textbf{Semantic conditioning beyond part count.} Geometry and the requested number of
  regions $K$ are fixed. The mismatched condition uses genuine same-$K$ annotation text with
  semantics that do not match the target grouping.
  We report raw area-weighted mIoU without post-processing, following \Secref{sec:setup}.}
  \label{tab:semantic_ablation}
  \small
  \setlength{\tabcolsep}{0.75em}
  \begin{tabular}{lcc}
    \toprule
    Prompt condition & mIoU $\uparrow$ & $\Delta$ \\
    \midrule
    Full semantic prompt                & 0.691 & -- \\
    $K$-only placeholders               & 0.373 & $-0.318$ \\
    Mismatched semantics, same $K$      & 0.493 & $-0.198$ \\
    \bottomrule
  \end{tabular}
\end{table}

Removing semantic content causes a large drop even though $K$ is unchanged. More importantly, a
genuine but mismatched same-$K$ prompt also reduces mIoU substantially. The predicted partition
therefore depends on the requested semantic content rather than acting only as a geometrically
plausible $K$-way decomposition. This test establishes sensitivity to semantic conditioning; it does
not evaluate a second, independently valid decomposition of the same object.

\subsection{Comparison with CubePart}
\label{sec:appendix_cubepart}
CubePart~\citep{zhu2026cubepart} conditions a generative model on an open-vocabulary part schema.
Its second stage can decompose an existing mesh, which makes it the closest schema-conditioned
method to our setting, and its code is released. A quantitative comparison under our metric is
nonetheless not available: CubePart synthesizes a separate mesh for each requested part rather than
labelling the faces of the input, so there is no per-face prediction to score without first
projecting its output back onto the input surface, a step of our choosing rather than theirs.
\Figref{fig:cubepart} therefore compares the two visually, on the same watertight mesh and with
semantically matched part names. Each method receives prompts in its required format, at both
coarse and fine granularities.

CubePart represents each requested item with its own latent and generates its geometry separately.
The number of generated latent groups therefore grows with the schema, and the resulting part meshes
must jointly reconstruct the input object. In these examples, that reconstruction becomes less stable
as the requested decomposition becomes more detailed. On the second object the coarse schema names \emph{both arms} and
\emph{both legs} as separate parts, yet CubePart's generated leg surface absorbs the arms, and at the
fine schema the right hand comes back as a fragment. The third object shows the same trend at the
scale our benchmark asks for: the coarse request returns a clean decomposition, while the detailed
request produces heavily fragmented part geometry despite passing CubePart's own validity check.

We treat these three objects as case studies rather than an estimate of CubePart's overall failure
rate. The displayed outputs show missing surface regions, geometry extending beyond the input shape,
and parts breaking into disconnected fragments. Our formulation avoids this geometry-generation
failure mode by retaining the input surface and predicting only its face labels.

\definecolor{cpgreen}{HTML}{1ADD99}
\definecolor{cpgrey}{HTML}{9A9A9A}
\definecolor{cppurple}{HTML}{BD96DC}

\newlength{\cpimsize}
\setlength{\cpimsize}{4.0cm}

\newcommand{\cpim}[3]{\includegraphics[trim=#3,clip,width=\cpimsize]{figures/12_cubepart/#1/#2.png}}
\newcommand{\cpbp}{{\color{black!12}\rule{\cpimsize}{\cpimsize}}}
\newcommand{\cphdr}[1]{\small #1}
\newcommand{\cprow}[2]{%
  \cpim{#1}{cp_coarse}{#2} & \cpim{#1}{cp_fine}{#2} &
  \cpim{#1}{ours_coarse}{#2} & \cpim{#1}{ours_fine}{#2} \\}
\newcommand{\cprowsplit}[3]{%
  \cpim{#1}{cp_coarse}{#2} & \cpim{#1}{cp_fine}{#3} &
  \cpim{#1}{ours_coarse}{#2} & \cpim{#1}{ours_fine}{#3} \\}
\newcommand{\cprowbp}{\cpbp & \cpbp & \cpbp & \cpbp \\}

\begin{figure*}[h]
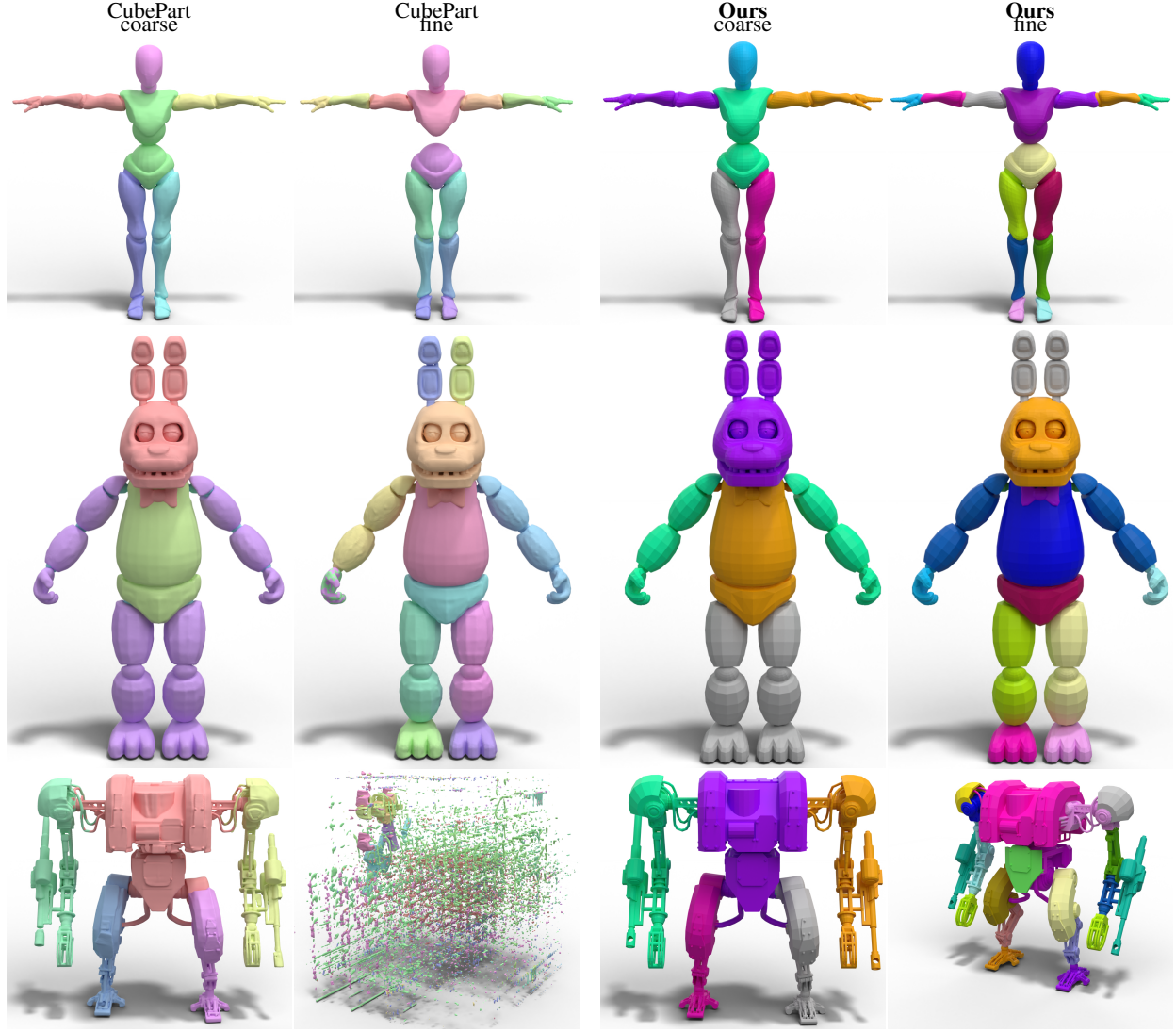

  \centering
  \setlength{\tabcolsep}{0.5pt}
  \renewcommand{\arraystretch}{0.0}
  \begin{tabular}{*{2}{c}@{\hspace{0.30cm}}*{2}{c}}
    \cphdr{CubePart} & \cphdr{CubePart} &
      \cphdr{\textbf{Ours}} & \cphdr{\textbf{Ours}} \\[-1pt]
    \cphdr{coarse} & \cphdr{fine} &
      \cphdr{coarse} & \cphdr{fine} \\[3pt]
    \cprow{3a538bf5ab9eff983c0d9d6e20796c232cd1cc4ccb8cb2865b6300d09ad4e9cf}{192 189 192 187}
    \cprow{8d4357a4152e421884823d3c2d3da14a}{350 198 396 186}
    \cprowsplit{00dcd724201ea40fd4475964332486b0}{131 202 202 225}{0 6 65 180}
  \end{tabular}
  \caption{\textbf{Following the prompt on an existing mesh.} Both methods receive the same watertight meshes and part names for coarse and fine decompositions. CubePart allocates and generates an independent latent geometry for each requested item, whereas our method retains the input surface and predicts its face labels. The displayed cases illustrate the resulting differences in partition and surface preservation.
\emph{Top:} At the fine level, CubePart fails to generate the waist, leaving the chest and pelvis disconnected; in contrast, our method continuously labels the original surface.
\emph{Middle:} The coarse prompt requests \emph{both arms} and \emph{both legs} as separate items, but CubePart erroneously merges them (\textcolor{cppurple}{\textbf{purple}}). Our method correctly separates the arms (\textcolor{cpgreen}{\textbf{green}}) and legs (\textcolor{cpgrey}{\textbf{grey}}). At the fine level, CubePart generates the right hand as fragmented specks.
\emph{Bottom:} CubePart handles the coarse request successfully but its fine-grained output (armour plates, brackets, hoses) becomes a cloud of fragments. Our method instead assigns labels on the unchanged input surface, yielding a complete partition without post-processing.}
  \label{fig:cubepart}
\end{figure*}

\subsection{Broader impact}
\label{sec:appendix_impact}

\paragraph{Positive impact.}
An explicit part decomposition is what part-aware workflows need: localized editing, rigging and
retopology, and semantic reuse of an asset's components. Because the control signal is text, a user
obtains one by naming the parts, and a frozen vision-language model can write the prompt from
renderings, which puts the capability within reach for education, indie development and rapid
prototyping. We label the faces of the input mesh to obtain a part decomposition of the user's asset.

\paragraph{Negative impact and mitigation.}
Like any generative 3D model, \Ours can be misused, for instance to decompose and repurpose a
copyrighted asset without attribution. It also inherits the biases of its training data, which
over-represents common object categories and the design conventions of the artists who built them,
so both its part vocabulary and its notion of what counts as one part are skewed towards those
categories. We train only on license-compliant data, we recommend that downstream users attribute
reused assets and check outputs for intellectual-property concerns before deployment, and we do not
intend the method for identity-related or surveillance applications.

\end{document}